\documentclass[conference]{IEEEtran}
\IEEEoverridecommandlockouts
\usepackage{multirow}
\usepackage{graphicx}
\usepackage{cite}
\usepackage{amsmath,amssymb,amsfonts}
\usepackage{algorithmic}
\usepackage{graphicx}
\usepackage{textcomp}
\usepackage{xcolor}
\def\BibTeX{{\rm B\kern-.05em{\sc i\kern-.025em b}\kern-.08em
    T\kern-.1667em\lower.7ex\hbox{E}\kern-.125emX}}
\begin{document}
\title{Automatic Wire-Harness Color Sequence Detector\\
{\footnotesize \textsuperscript{}}
\thanks{}
}

\author{\IEEEauthorblockN{1\textsuperscript{st} Indiwara Nanayakkara}
\IEEEauthorblockA{\textit{INSIGHT I Q LABS} \\
Katunayake, Sri Lanka \\
indiwarananayakkara1@gmail.com}
\and
\IEEEauthorblockN{2\textsuperscript{nd} Dehan Jayawickrama}
\IEEEauthorblockA{\textit{INSIGHT I Q LABS} \\
Katunayake, Sri Lanka \\
dehanjayawickrama@gmail.com}
\and
\IEEEauthorblockN{3\textsuperscript{rd} Mervyn Parakrama B. Ekanayake}
\IEEEauthorblockA{\textit{Department of Electrical and Electronics Engineering} \\
\textit{Faculty of Engineering, University of Peradeniya}\\
Peradeniya, Sri Lanka \\
mpb.ekanayake@gmail.com}
}

\maketitle

\IEEEpubidadjcol
\begin{abstract}

Wire harness inspection process remains a labor-intensive process prone to errors in the modern Electronics Manufacturing Services (EMS) industry.  This paper introduces a semiautomated machine vision system capable of verifying correct wire positioning, correctness of the connector polarity and correctness of color sequences for both linear and circular wire harness configurations. Five industrial standard CMOS cameras are integrated into a modularized mechanical framework in the physical structure of the solution and a HSV and RGB color domain value comparison based color sequence classifier is used in the operation. For each harness batch, a user can train the system using at least five reference samples; the trained file is stored and reused for similar harness types. The Solution is deployed at GPV Lanka Pvt. Ltd. (Fig. 2) and the system achieved 100\% detection accuracy and reduced inspection time by 44\% compared to manual methods. Additional features include user management, adjustable lighting, session data storage, and secure login. Results of this product usage in the real world situation demonstrate that this approach delivers reliable and efficient inspection capabilities. \cite{1,2,3}.

\end{abstract}
\section{Introduction}

Wire Harness is a structured bundle of wires, cables and connectors that efficiently transmit electrical power and signals within a system.
Wire harnesses are critical in electronic assemblies, where their functionality directly impacts product reliability and safety. Producing wire harnesses is one of the operations in the EMS industry.  When wire harnesses are produced in the EMS industry manufacturers follow several steps based on their client requirements. Some of these steps are value added and some are non-value added. Clients are charged only for the value added steps in the manufacturing process and non-value added steps are not being used in the charges. In order to produce an output that is desired by a customer non-value added steps are essential. One of the major concerns in the EMS industry is to minimize the time utilized in the non-value added steps.  Among the value added EMS operations in wire harnesses manufacturing that must meet strict quality requirements are wire cutting, crimping, and insertion into connectors. 
After a wire harness is produced it is checked for its quality with the customer specifications. Wire harnesses are inspected as batches  in  hundreds and even thousands of quantities. Conventional quality inspection process is performed manually by a majority of EMS service providers.  Manual inspection methods are time consuming, unreliable and inefficient, particularly in detecting subtle defects like incorrect wire sequences or connector misplacement \cite{4,5}.
Current automation solutions are primarily tailored to linear harnesses only, with no adaptability to circular wire harness configurations.\cite{6,7}. This paper presents a machine vision system, which automates inspection for both linear and circular wire harnesses (Fig. 1) using five industrial cameras and a lightweight neural network architecture based classification techniques. Before the machine was deployed in a production process it went through rigorous quality testing (Table 1), showing promising results in accuracy and efficiency \cite{8}

Under this project scope the authors identified three types of wire harnesses.
\begin{enumerate}
    \item Single row linear connector wire harness (Fig. 4)
    \item Double row linear connector wire harness (Fig. 5)
    \item Circular shaped wire harness (Fig. 6)
\end{enumerate}

\begin{figure}[htbp]
    \centering
    \includegraphics[width=0.35\textwidth]{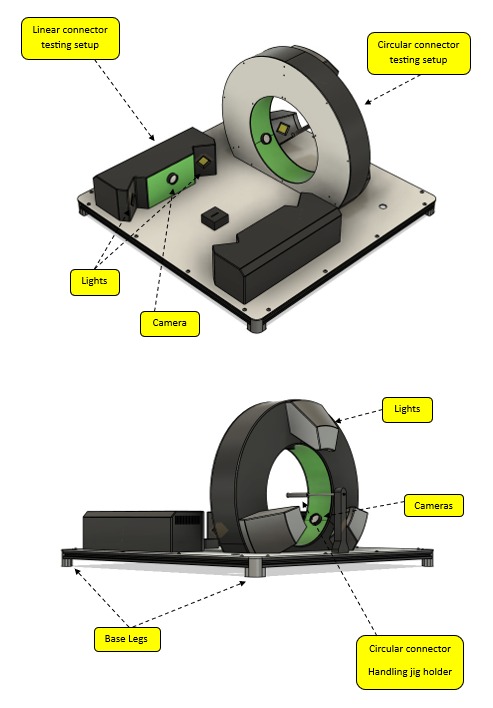}
    \caption{Machine Setup Diagram}
    \label{fig:image1}
\end{figure}

\begin{figure}[htbp]
    \centering
    \includegraphics[width=0.35\textwidth]{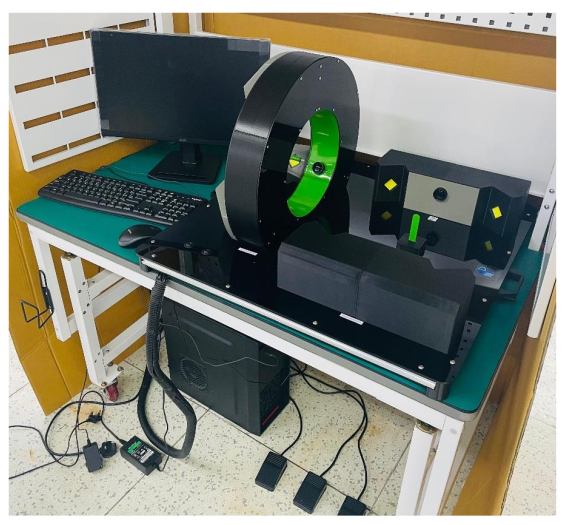}
    \caption{Production ready implementation}
    \label{fig:image2}
\end{figure}

\begin{figure}[htbp]
    \centering
    \includegraphics[width=0.33\textwidth]{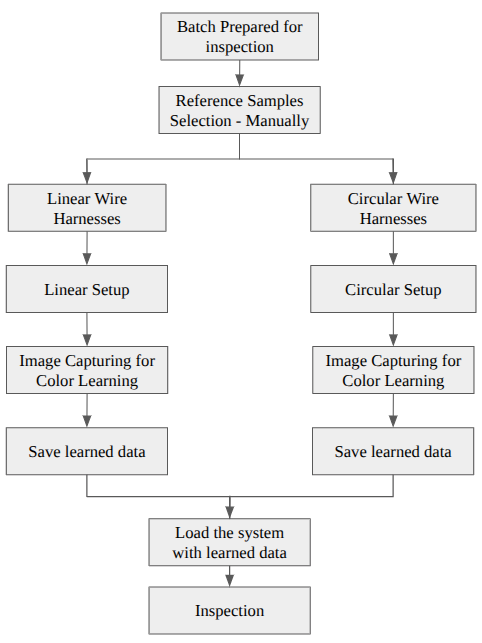}
    \caption{System Workflow Diagram}
    \label{fig:image3}
\end{figure}
\section{System Overview}

The system workflow consists of the following stages in an EMS production environment:
\begin{enumerate}
    \item A batch of wire harnesses is prepared for an inspection.
    \item The supervisor selects a minimum of five correct samples (training set); optionally, more samples (6, 7, or 8) can be used for better generalization.
    \item The system captures images of the linear connectors using the "Linear connector testing set-up" and it captures the images of circular connectors using the "Circular circuit testing set-up" to perform RGB and HSV color domain value based comparison.
    \item The learning file is stored and can be reused for the same harness type in future production runs.
    \item Subsequent harnesses are inspected against the learned data for color sequence and connector polarity.
\end{enumerate}

\begin{figure}[htbp]
    \centering
    \includegraphics[width=0.35\textwidth]{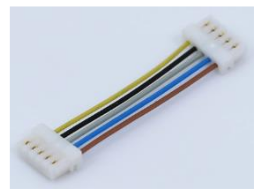}
    \caption{Linear Single Row Connector}
    \label{fig:image4}
\end{figure}

\begin{figure}[htbp]
    \centering
    \includegraphics[width=0.35\textwidth]{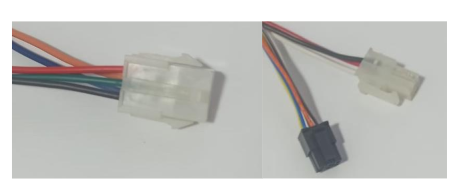}
    \caption{Linear Double Row Connector}
    \label{fig:image5}
\end{figure}

\begin{figure}[htbp]
    \centering
    \includegraphics[width=0.35\textwidth]{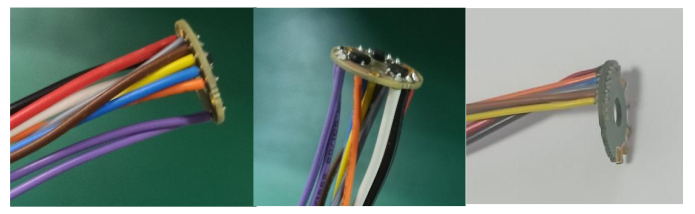}
    \caption{Circular Connectors}
    \label{fig:image6}
\end{figure}

\begin{figure}[htbp]
    \centering
    \includegraphics[width=0.20\textwidth]{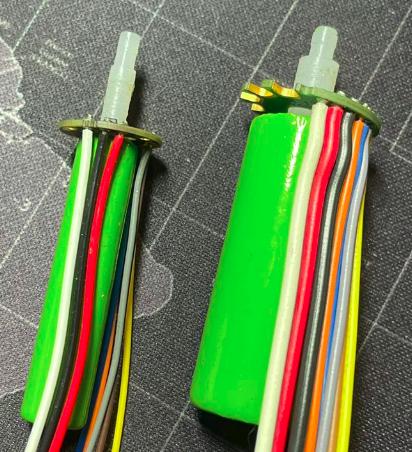}
    \caption{Circular connector jigs}
    \label{fig:image7}
\end{figure}

\section{Proposed Methodology}

\subsection{Hardware Setup}

In order to test linear connectors the linear connector testing setup with two cameras is used. The setup can test both linear single row and double row connector types shown in Fig. 8 and Fig. 9 respectively. For the circular connector testing the 3 cameras setup is used which captures all the wires in a circular connector attached to a jig (Fig. 6 and Fig. 7) in 120 degree angles (Fig. 10).

\subsection{Camera Configuration}

Gray cards and color charts were used to calibrate white balance, saturation and brightness based on reference RGB values obtained under controlled lighting conditions. During deployment the camera parameters were adjusted to match these references to ensure consistent color accuracy and minimal camera drift.

\subsection{Application Packaging and Deployment}

The application along with the machine learning model has been deployed offline as a Debian package in an Ubuntu environment. 

\subsection{Training and Testing with the System}

During the single row linear connector wire harness training, average pixel-wise color values of the 5 images are taken into account in both RGB and HSV color spaces. Only a single camera is used in this scenario (Fig. 8). These average values are stored in a trained file for that particular wire harness type. 
In the testing phase the same trained file is used with the average color values and they are compared with the color values of the testing sample's color values against their Mean Square Error (MSE) values for each pixel. This process is carried out for both double row linear connector based wire harnesses and circular connector based wire harness training and testing. This process is further explained in section IV. 

\begin{figure}[htbp]
    \centering
    \includegraphics[width=0.35\textwidth]{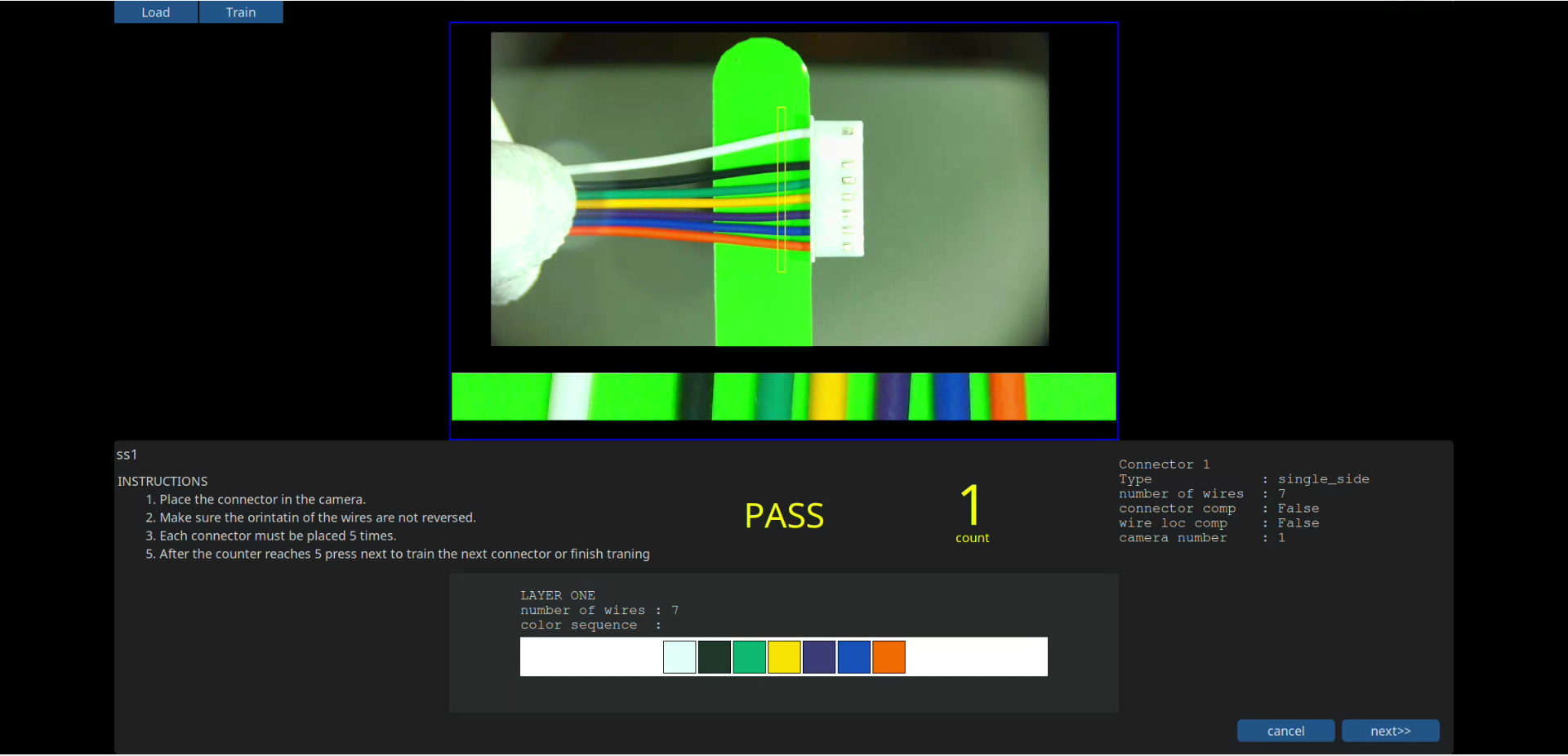}
    \caption{Test UI for single layer connector.}
    \label{fig:image8}
\end{figure}

\begin{figure}[htbp]
    \centering
    \includegraphics[width=0.35\textwidth]{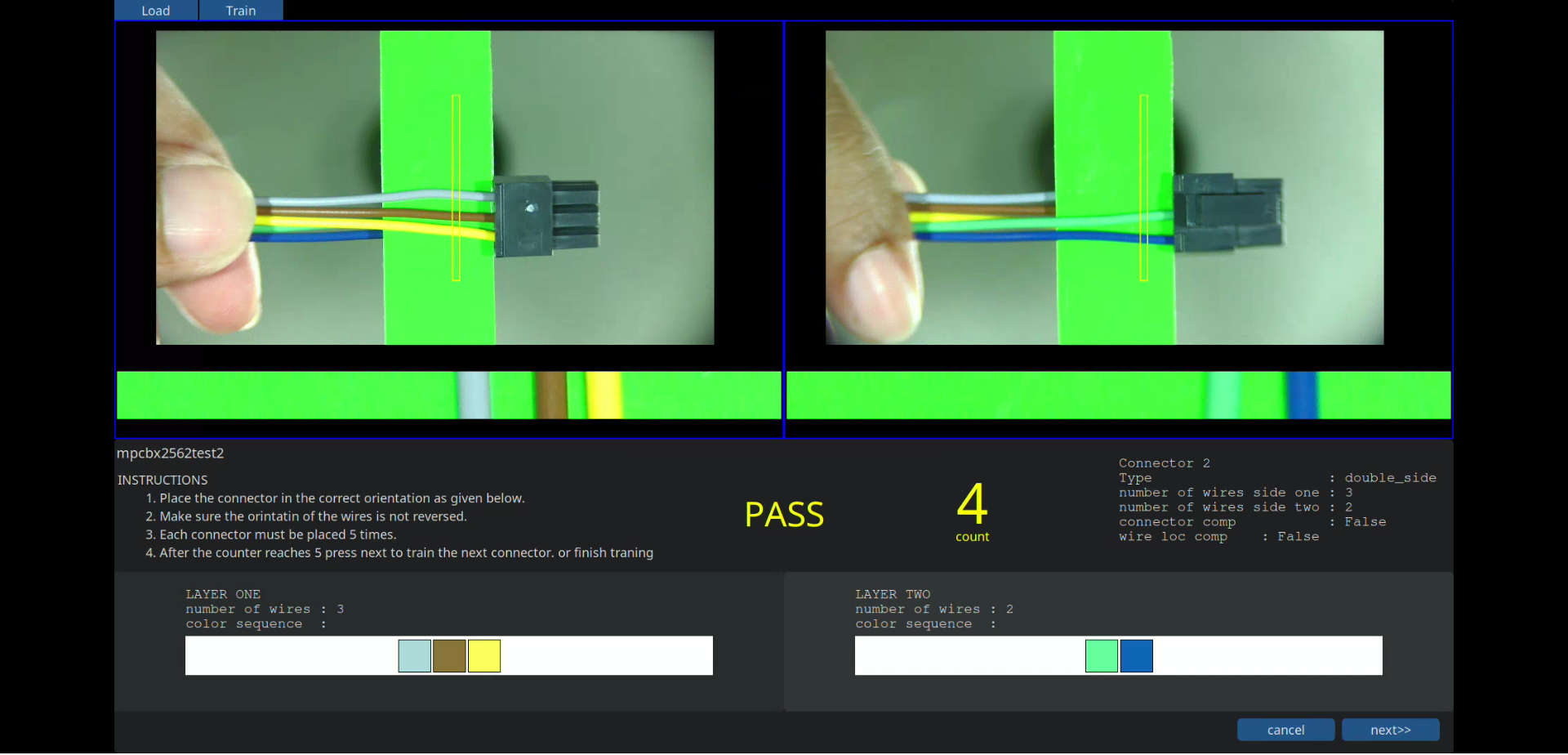}
    \caption{Linear Double Side Connector Testing}
    \label{fig:image9}
\end{figure}

\begin{figure}[htbp]
    \centering
    \includegraphics[width=0.35\textwidth]{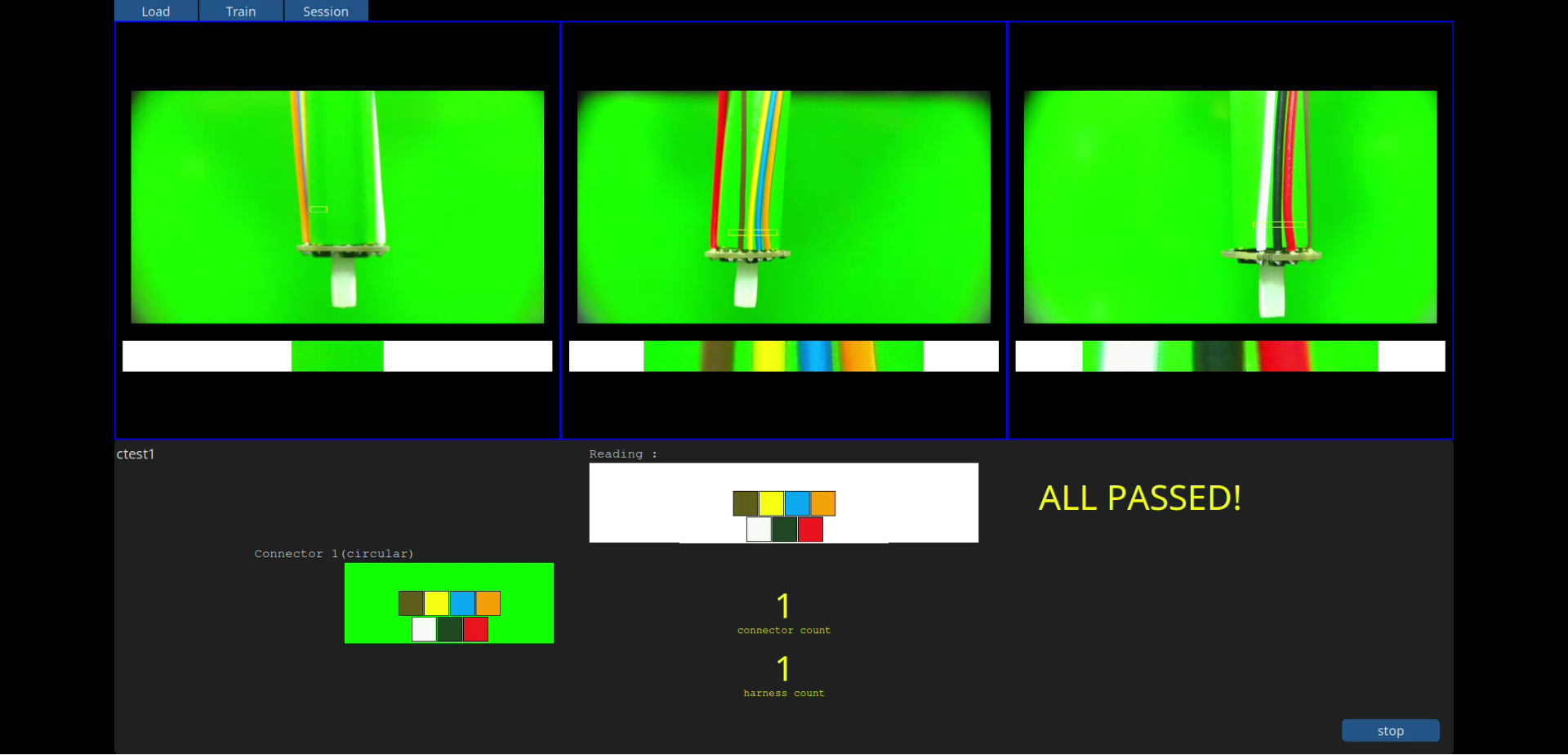}
    \caption{Circular Connector Testing}
    \label{fig:image10}
\end{figure}

\section{Color Comparison Algorithm}
To perform an accurate color comparison, the first step involves identifying different bounding boxes within the image. Each bounding box corresponds to a specific wire and defines the region from which the pixel values will be extracted. These regions isolate each wire for individual color analysis. The detection and comparison process is outlined in Fig. 17.

\subsection{Step 1: Crop the Region of Interest}
In this step, the frame captured from the camera feed is cropped using a predefined Region of Interest (ROI). This ROI is specified during the training process and remains fixed during testing. The portion of the image within this ROI is extracted and referred to as the \textbf{cropped frame}, (Fig. 11) which is then used in subsequent steps.

\begin{figure}[htbp]
    \centering
    \includegraphics[width=0.35\textwidth]{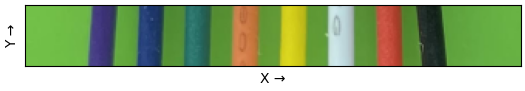}
    \caption{Cropped Frame}
    \label{fig:image11}
\end{figure}

\subsection{Step 2: HSV-Based Color Segmentation}
\begin{figure}[htbp]
    \centering
    \includegraphics[width=0.35\textwidth]{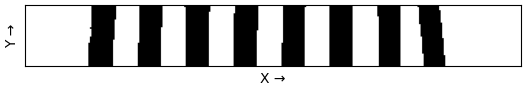}
    \caption{Background Mask and Wires Separation}
    \label{fig:image12}
\end{figure}

To isolate the background from the wires, HSV-based thresholding is applied to the cropped frame from Step 1. Given that the background has a predefined color ("Green" in this use case), this process produces a binary mask that highlights the background pixels as 255 and the wire pixels as 0, as shown in Fig.~\ref{fig:image12}. This mask will be known as the \textbf{background mask}.

\subsection{Step 3: Identify Start and End Points of the Bounding Box}
To identify bounding box endpoints, four horizontal lines are sampled along the vertical ($y$) axis at positions: $y = 0$, $y = y_\text{max}$, $y = 0.1 \times y_\text{max}$, and $y = 0.9 \times y_\text{max}$, where $y_\text{max}$ is the height of the cropped frame. Lines at $y = 0$ and $y = 0.1 \times y_\text{max}$ detect lower endpoints, while $y = y_\text{max}$ and $y = 0.9 \times y_\text{max}$ detect upper endpoints. Fallback lines ($0.1 \times y_\text{max}$ and $0.9 \times y_\text{max}$) are used only if primary lines fail.

Endpoint detection along each line is done by computing the difference between the background mask and a version shifted by one pixel along the $x$-axis. Wire boundaries yield a difference of 255 (background = 255, wire = 0). The expected number of endpoints is known based on the number of wires (e.g., 8 wires result in 16 endpoints along a single line - Fig. 13). If the detected number of endpoints differs from this expected value, fallback lines are used to maintain robustness. If both the upper and lower wire endpoints are successfully detected, Step~4 is skipped and the process proceeds directly to Step~5.

\begin{figure}[htbp]
    \centering
    \includegraphics[width=0.35\textwidth]{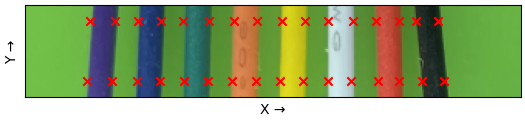}
    \caption{Identified start and end points}
    \label{fig:image13}
\end{figure}

\subsection{Step 4: Compute Gradient Mask}

When wires are closely spaced, the background mask may fail due to the absence of visible pixels between them. To address this, a gradient-based method is used to detect wire boundaries, overcoming issues caused by printed text, reflections, and merged edges. \cite{12} The process includes the following steps:

\subsubsection{Gradient Calculation}
Compute the x-direction gradient of the grayscale cropped image of size $(n \times m)$, resulting in a gradient array of shape $(n \times (m-1))$.

\subsubsection{Thresholding}
Threshold the gradient to suppress weak or noisy edges. This produces a binary gradient map of the same size.

\subsubsection{Vertical Summation}
Sum the thresholded gradient along the y-axis to produce a $(1 \times (m-1))$ array, highlighting columns with strong horizontal transitions—potential wire boundaries.

\subsubsection{Segment Identification}
Identify continuous segments where the summed values exceed a defined threshold. (Fig. 14) Each segment is treated as a candidate region containing a wire boundary. Additional conditions like length constraints are used to identify the exact candidates for wire boundaries.

\begin{figure}[htbp]
    \centering
    \includegraphics[width=0.49\textwidth]{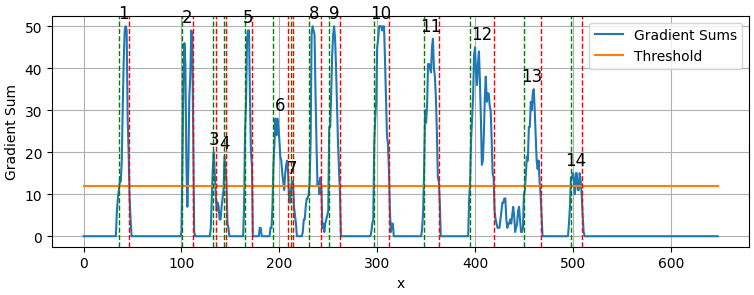}
    \caption{Gradient sum plot with identified segments}
    \label{fig:image14}
\end{figure}

\subsubsection{Line Matching Using Predefined Templates}

Each segment identified in the previous step is further evaluated using binary line templates to confirm and localize the wire boundary. The process is as follows:

\begin{itemize}
    \item For each segment, the x-coordinate corresponding to the maximum value in the vertically summed gradient array is selected as the center point $x_1$.
    \item A crop of size $(n \times m_\text{template})$ is extracted from the thresholded gradient map, centered at $x_1$. This crop spans horizontally from $x_1 - \frac{m_\text{template}}{2}$ to $x_1 + \frac{m_\text{template}}{2}$.
    \item A set of predefined binary line template each of shape $(n \times m_\text{template})$ is used. Each template contains a centered line of a specific slope, represented by 1s along the line and 0s elsewhere. (Fig. 15)
    \item Each template is bitwise ANDed with the cropped segment. The overlap is calculated as the number of matching 1s divided by the total number of 1s in the template.
    \item If the overlap exceeds 90\%, the segment is accepted as a valid boundary, and the best-matching template is assigned to that segment.
\end{itemize}

This is repeated for all segments. For each accepted segment, we retain the center coordinate ($x_1$) and the matched template.

\begin{figure}[htbp]
    \centering
    \includegraphics[width=0.35\textwidth]{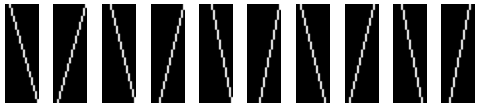}
    \caption{Some Predefined Line Templates}
    \label{fig:image15}
\end{figure}

\subsubsection{Gradient Mask Construction}
A zero-initialized binary array of the same shape as the \textbf{cropped frame} is created. All matched templates are placed at their respective center coordinates, resulting in a binary image with lines representing estimated wire boundaries. This output is referred to as the \textbf{gradient mask}. (Fig. 16)

\subsubsection{Final Mask Generation}
A bitwise AND is performed between the \textbf{background mask} and the \textbf{gradient mask}. Since the background mask contains 1s in non-wire regions and 0s at wire locations, this operation suppresses regions not confirmed by both masks. The updated mask is then fed back to Step~3 for improved bounding box endpoint detection \cite{13}. (Fig. 16)

\begin{figure}[htbp]
    \centering
    \includegraphics[width=0.35\textwidth]{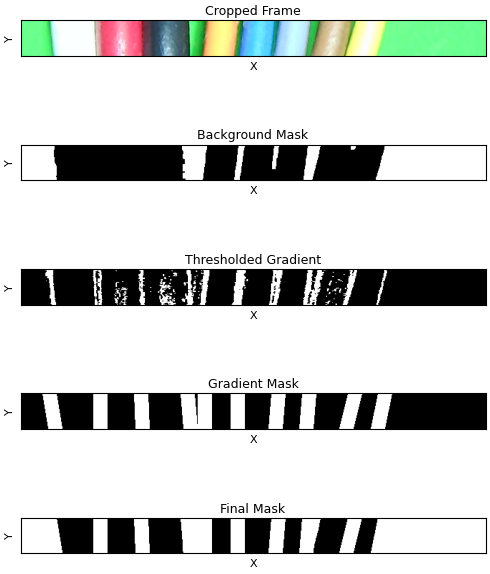}
    \caption{Gradient mask generation steps}
    \label{fig:image16}
\end{figure}

\subsection{Step 5: Generate Bounding Boxes}
From Step~3, four endpoints are identified for each wire (see Fig. ~\ref{fig:image13}). Vertical ($y$) endpoints are fixed based on predefined lines, while horizontal ($x$) endpoints are refined by selecting the maximum of the left pair and the minimum of the right pair. This ensures each bounding box tightly encloses the wire while excluding background and adjacent wires.

\subsection{Step 6: Color Comparison}
Once bounding boxes are defined, colors within each box are compared to reference colors from the training phase using Mean Squared Error (MSE) in both RGB and HSV color spaces.

\begin{figure}[htbp]
    \centering
    \includegraphics[width=0.5\textwidth]{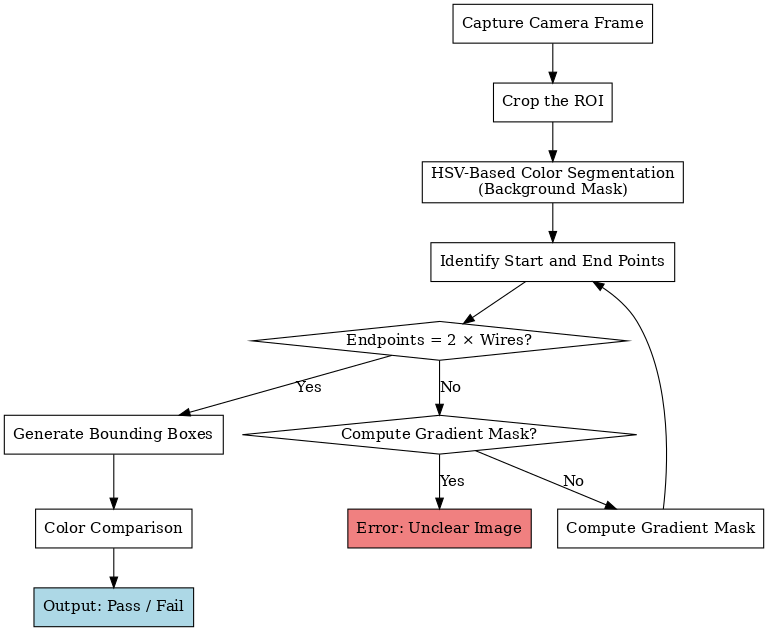}
    \caption{Wire Harness Inspection and Color Comparison Workflow}
    \label{fig:image17}
\end{figure}

\section{Connector Orientation Validation for Wire Sequence Verification}
While verifying the color sequence of wires connected to a connector is important, it is possible for the wire order to appear correct when the connector itself is inserted in the reverse orientation (Fig. 18). In such cases, although the color sequence remains valid, the physical orientation of the connector may be incorrect. To address this issue, a connector comparison algorithm is introduced. Two different strategies are employed, depending on whether the two sides of the connector are visually distinguishable.
\begin{figure}[htbp]
    \centering
    \includegraphics[width=0.33\textwidth]{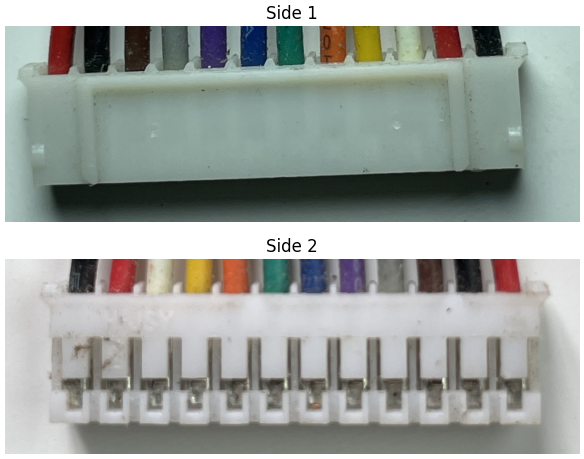}
    \caption{Two sides of the same connector.}
    \label{fig:image18}
\end{figure}
\subsection{Visually Distinct Connectors}
For connectors with identifiable sides (e.g., a notch on one side), orientation is verified using visual feature comparison. A reference image of the correctly oriented connector is passed through a truncated MobileNetV3 to extract a feature vector \cite{14,15}. During inference, the test image is processed similarly, and cosine similarity is computed. A high score confirms correct orientation.

\subsection{Symmetrical Connectors}
When connector sides are visually indistinguishable, a colored marker (typically green - Fig. 19) is manually placed on one side during assembly. During inspection, the system checks for this marker within the connector region. Its presence confirms the correct orientation.

This two-step strategy ensures both color sequence and connector orientation are validated, even for visually symmetrical connector types.

\begin{figure}[htbp]
    \centering
    \includegraphics[width=0.35\textwidth]{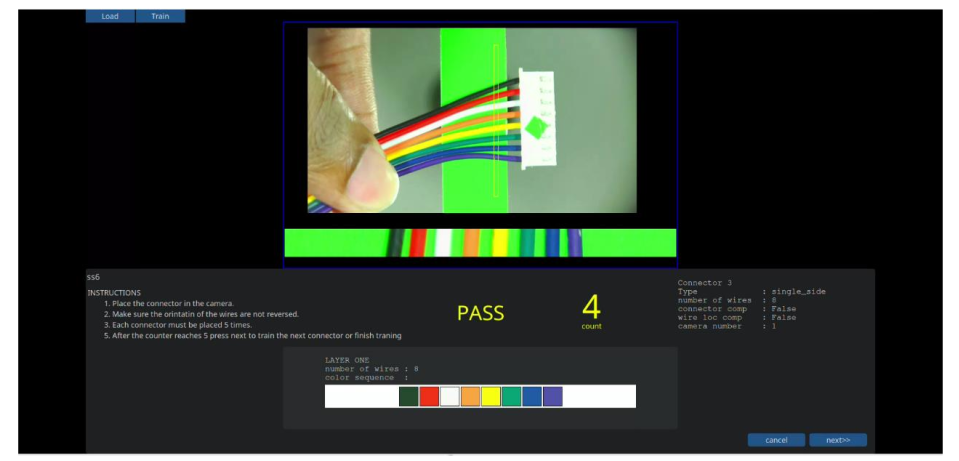}
    \caption{Symmetrical Connector Side Identification}
    \label{fig:image19}
\end{figure}

\section{Results}

When the machine was setup at the production facility a User Acceptance Test (UAT) was conducted with all the functions being tested indicated in Table 1. 

After all the machine functionalities have been tested a production ready wire harness (Fig. 20) with a quantity of 500 samples were tested. Time was measured for both manual inspection process and machine inspection process. 

\begin{itemize}
    \item Average time for a worker to conduct the quality inspection process without using the machine : 2.5 hours (18 seconds / harness )
    \item Average time for a trained worker on the machine to conduct the quality inspection process with the machine : 1.4 hours (10 seconds / harness )
\end{itemize}

In the implemented system an operator is stationed at the workstation to handle ambiguous cases. When the classifier is unable to identify the harness correctly, the user interface displays the message "Image not clear", prompting the operator to perform a manual inspection on the harness. This human-in-the-loop fallback prevents uncertain outputs from propagating through the workflow and thereby helps mitigate errors while preserving overall system reliability.

\begin{table}[]
\centering
\caption{Tests conducted on the machine before production usage}
\resizebox{\columnwidth}{!}{%
\begin{tabular}{|cll|}
\hline
\multicolumn{1}{|c|}{\multirow{14}{*}{\begin{tabular}[c]{@{}c@{}}Single Row \\ Training/Testing\end{tabular}}} &
  \multicolumn{1}{l|}{single row single connector - cam 1} &
  pass \\ \cline{2-3} 
\multicolumn{1}{|c|}{} &
  \multicolumn{1}{l|}{single row single connector - cam 2} &
  pass \\ \cline{2-3} 
\multicolumn{1}{|c|}{} &
  \multicolumn{1}{l|}{single row multiple connectors - cam 1} &
  pass \\ \cline{2-3} 
\multicolumn{1}{|c|}{} &
  \multicolumn{1}{l|}{single row multiple connectors - cam 2} &
  pass \\ \cline{2-3} 
\multicolumn{1}{|c|}{} &
  \multicolumn{1}{l|}{single row multiple connectors cam 1 and cam 2} &
  pass \\ \cline{2-3} 
\multicolumn{1}{|c|}{} &
  \multicolumn{1}{l|}{single row multiple connectors placement checking cam 1} &
  pass \\ \cline{2-3} 
\multicolumn{1}{|c|}{} &
  \multicolumn{1}{l|}{single row multiple connectors placement checking cam 2} &
  pass \\ \cline{2-3} 
\multicolumn{1}{|c|}{} &
  \multicolumn{1}{l|}{\begin{tabular}[c]{@{}l@{}}single row multiple connectors placement checking cam 1\\  and cam 2\end{tabular}} &
  pass \\ \cline{2-3} 
\multicolumn{1}{|c|}{} &
  \multicolumn{1}{l|}{single row multiple connectors polarity checking cam 1} &
  pass \\ \cline{2-3} 
\multicolumn{1}{|c|}{} &
  \multicolumn{1}{l|}{single row multiple connectors polarity checking cam 2} &
  pass \\ \cline{2-3} 
\multicolumn{1}{|c|}{} &
  \multicolumn{1}{l|}{\begin{tabular}[c]{@{}l@{}}single row multiple connectors polarity checking cam 1 \\ and cam 2\end{tabular}} &
  pass \\ \cline{2-3} 
\multicolumn{1}{|c|}{} &
  \multicolumn{1}{l|}{\begin{tabular}[c]{@{}l@{}}single row multiple connectors polarity checking and \\ placement checking cam 1\end{tabular}} &
  pass \\ \cline{2-3} 
\multicolumn{1}{|c|}{} &
  \multicolumn{1}{l|}{\begin{tabular}[c]{@{}l@{}}single row multiple connectors polarity checking and\\ placement checking cam 2\end{tabular}} &
  pass \\ \cline{2-3} 
\multicolumn{1}{|c|}{} &
  \multicolumn{1}{l|}{\begin{tabular}[c]{@{}l@{}}single row multiple connectors polarity checking and\\ placement checking cam 1 and cam 2\end{tabular}} &
  pass \\ \hline
\multicolumn{3}{|l|}{} \\ \hline
\multicolumn{1}{|c|}{\multirow{7}{*}{\begin{tabular}[c]{@{}c@{}}Double Row\\ Training/Testing\end{tabular}}} &
  \multicolumn{1}{l|}{doubel row single connector} &
  pass \\ \cline{2-3} 
\multicolumn{1}{|c|}{} &
  \multicolumn{1}{l|}{double row multiple connectors} &
  pass \\ \cline{2-3} 
\multicolumn{1}{|c|}{} &
  \multicolumn{1}{l|}{double row and single row connectors} &
  pass \\ \cline{2-3} 
\multicolumn{1}{|c|}{} &
  \multicolumn{1}{l|}{double row placement checking} &
  pass \\ \cline{2-3} 
\multicolumn{1}{|c|}{} &
  \multicolumn{1}{l|}{\begin{tabular}[c]{@{}l@{}}double  polarity checking\\ i. both sides no sticker\\ ii.both sides green sticker\end{tabular}} &
  pass \\ \cline{2-3} 
\multicolumn{1}{|c|}{} &
  \multicolumn{1}{l|}{double side polarity checking and placement checking} &
  pass \\ \cline{2-3} 
\multicolumn{1}{|c|}{} &
  \multicolumn{1}{l|}{\begin{tabular}[c]{@{}l@{}}double side and single side connectors with random\\  placement and polarity checking\end{tabular}} &
  pass \\ \hline
\multicolumn{3}{|l|}{} \\ \hline
\multicolumn{1}{|c|}{\multirow{6}{*}{Circular}} &
  \multicolumn{1}{l|}{single set cam 1} &
  pass \\ \cline{2-3} 
\multicolumn{1}{|c|}{} &
  \multicolumn{1}{l|}{single set cam 2} &
  pass \\ \cline{2-3} 
\multicolumn{1}{|c|}{} &
  \multicolumn{1}{l|}{single set cam 3} &
  pass \\ \cline{2-3} 
\multicolumn{1}{|c|}{} &
  \multicolumn{1}{l|}{double set cam 1 and cam 2} &
  pass \\ \cline{2-3} 
\multicolumn{1}{|c|}{} &
  \multicolumn{1}{l|}{double set cam 2 and cam 3} &
  pass \\ \cline{2-3} 
\multicolumn{1}{|c|}{} &
  \multicolumn{1}{l|}{triple set all cams} &
  pass \\ \hline
\multicolumn{3}{|l|}{} \\ \hline
\multicolumn{1}{|c|}{\multirow{5}{*}{\begin{tabular}[c]{@{}c@{}}Combined \\ Testing/Training\end{tabular}}} &
  \multicolumn{1}{l|}{single row connector cam 1 with circular} &
  pass \\ \cline{2-3} 
\multicolumn{1}{|c|}{} &
  \multicolumn{1}{l|}{single row connector cam 2 with circular} &
  pass \\ \cline{2-3} 
\multicolumn{1}{|c|}{} &
  \multicolumn{1}{l|}{single row connectors cam 1 and cam 2 with circular} &
  pass \\ \cline{2-3} 
\multicolumn{1}{|c|}{} &
  \multicolumn{1}{l|}{double row connector with circular} &
  pass \\ \cline{2-3} 
\multicolumn{1}{|c|}{} &
  \multicolumn{1}{l|}{\begin{tabular}[c]{@{}l@{}}single row double row and circular connectors with\\  random placement and polarity checking\end{tabular}} &
  pass \\ \hline
\end{tabular}%
}
\label{tab:example} 
\end{table}

\begin{figure}[htbp]
    \centering
    \includegraphics[width=0.33\textwidth]{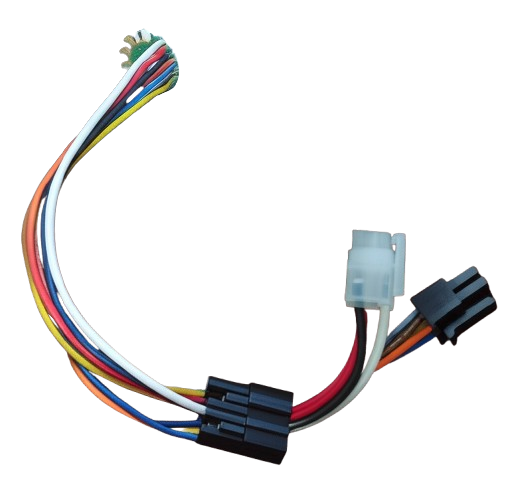}
    \caption{Wire Harness used in initial production testing - Single Row, Double Row and Circular Connectors are tested}
    \label{fig:image20}
\end{figure}

\section{System Limitations and Future Work}

Wires with spiral strips on different colors are unable to identified using this system. Current setup needs to be triggered by a human using a foot switch each time a wire harness is checked and the foot switch integration has been done based on the project requester requirements. However, the system can be improved with a conveyor system to fully automate the triggering mechanism in the future upgrades.

\section{Conclusion}

This work presents a practical machine vision system designed to improve the quality inspection of wire harnesses in EMS industry. Capable of handling both linear and circular configurations along with customization user features. Its deployment in a local factory in Sri Lanka demonstrated 100\% detection accuracy, a substantial efficiency increase and reliable detection of non-trivial anomalies like incorrect connector orientation \cite{9,10,11}. Future development will focus on broader harness types, improved retraining and integration with manufacturing data systems.

\bibliography{Ref}

\end{document}